\documentclass[reprint,
superscriptaddress,
amsmath,
amssymb,
aps,
twocols
]{revtex4-1} 
\usepackage{verbatim}
\usepackage{graphicx}
\usepackage{dcolumn}
\usepackage{bm}
\usepackage{mathtools}

\usepackage{float}

\usepackage{graphicx}
\usepackage{hyperref}
\usepackage{multirow}
\usepackage{hhline}                         
\usepackage[usenames, dvipsnames]{color}

\usepackage{amssymb, amsfonts}
\usepackage{booktabs}
\usepackage[ruled]{algorithm2e}
\usepackage{algcompatible}
\usepackage{cancel}

\newcolumntype{P}[1]{>{\centering\arraybackslash}p{#1}}

\newcommand{\ie}{i.e., }

\usepackage{lipsum}

\begin{document}
\renewcommand{\tableautorefname}{Tab.}
\renewcommand{\figureautorefname}{FIG.}
\renewcommand{\sectionautorefname}{Section}

\title{Coarse-Graining Hidden Representations:\\ Unsupervised Neuron Selection via Mapping Entropy}

\author{Margherita Mele}
\affiliation{Physics Department, University of Trento, via Sommarive, 14 I-38123 Trento, Italy}
\affiliation{INFN-TIFPA, Trento Institute for Fundamental Physics and Applications, I-38123 Trento, Italy}

\author{Andrea Castagna}
\affiliation{Physics Department, University of Trento, via Sommarive, 14 I-38123 Trento, Italy}

\author{Roberto Menichetti}
\affiliation{Physics Department, University of Trento, via Sommarive, 14 I-38123 Trento, Italy}
\affiliation{INFN-TIFPA, Trento Institute for Fundamental Physics and Applications, I-38123 Trento, Italy}

\author{Raffaello Potestio}
\email{raffaello.potestio@unitn.it}
\affiliation{Physics Department, University of Trento, via Sommarive, 14 I-38123 Trento, Italy}
\affiliation{INFN-TIFPA, Trento Institute for Fundamental Physics and Applications, I-38123 Trento, Italy}

\author{Alessandro Ingrosso}
\email{alessandro.ingrosso@donders.ru.nl}
\affiliation{Donders Centre for Neuroscience, Radboud University, Nijmegen, The Netherlands}

\date{\today}

\begin{abstract}
Overparameterized neural networks carry far more hidden units than a task nominally requires, raising the question of which neurons are essential and whether that distinction is legible in the representation itself, without labels or gradients. We cast neuron selection as the problem of coarse-graining the hidden layer by retaining a subset of its neurons, and score each putative selection by the mapping entropy (ME). This quantity measures the loss of discriminatory power inherent in discarding part of the network neurons, and the selection that minimises the ME is taken as particularly informative. This criterion is fully unsupervised, in that it depends only on hidden-activation statistics. In teacher–student networks, ME optimisation recovers the minimal teacher-consistent representation and retains extra units in proportion to the hidden layer's residual variability; in a non-linear Gaussian process task, it selects coherent functional-class mappings whose preferred class shifts across training. On this task and on translation-augmented MNIST, ME-selected subnetworks outperform random subsets of equal size, most clearly under strong compression — linking configurational distinguishability to predictive performance.
\end{abstract}

\maketitle

\section{Introduction}

Modern neural networks have achieved remarkable empirical success, even though the principles through which learning, generalisation, and internal organisation emerge are not yet fully understood.
These networks are generally highly overparameterized, in the sense that in order to achieve a high level of generalisation the system has to entail far more internal degrees of freedom than they are nominally required to encode the desired target input-output relation \citep{goodfellow2016deep,zdeborova2020understanding,mehta2019high,engel2001statistical}. The specific regime in which a network operates strongly shapes its resulting representations. In the infinite-width limit, training dynamics linearise around initialisation and the network is described by a fixed kernel, the neural tangent kernel \cite{jacot2018,lee2019}; in this \emph{lazy} regime the internal features barely move, and the hidden representation is essentially inherited from the random initial weights \cite{chizat2019}. The complementary \emph{feature-learning} or \emph{rich} regime is the one in which hidden units reorganise substantially during training, developing structured, data-adapted representations~\cite{mei2018,geiger2020,yang2021}.

Widely regarded as a central ingredient of the optimisation dynamics and empirical success of neural networks, overparameterization raises fundamental questions regarding the nature of feature learning in deep networks. How is task-relevant computation distributed across the hidden units? Can we identify specific neurons that are essential to preserve predictive performance and others that, instead, play a redundant or secondary role in a learned representation? These questions also bear on two distinct but closely related concerns in deep learning: efficiency and interpretability.

Efficiency pertains to identifying the neurons that carry out most of the task-relevant computation, which is central to pruning and model compression---reducing memory and compute while preserving predictive performance \cite{lecun1990second,hassibi1993optimal,han2016deep,blalock2020state,hinton2015distilling}. Strategies to reduce network complexity while retaining accuracy are essential for deployment in resource-constrained settings, and they bear on transferability, since a useful compressed representation should remain adaptable under fine-tuning on downstream tasks \cite{yosinski2014transferable,iofinova2022well}. Compression is especially
valuable when labelled data are scarce, where one wishes to keep only the most robust and transferable features while shedding unnecessary capacity.

The same motivations drive the interpretability agenda. Here one asks whether individual hidden units have identifiable functional roles, how strongly information is localised versus distributed, and whether dominant features can be disentangled within the learned representation~\cite{bau2017network,morcos2018importance,hooker2019benchmark}. This question has been brought into sharp focus by recent work in \emph{mechanistic interpretability}, which seeks to reverse-engineer trained networks (transformers in particular) into human-understandable computational units. This literature has shown that
individual neurons are frequently \emph{polysemantic}, responding to superpositions of unrelated features, and has proposed that networks pack more features than they have neurons through \emph{superposition} \cite{elhage2022}; sparse dictionary methods such as sparse autoencoders have since been used to extract more monosemantic, interpretable directions from the residual stream of large language models~\cite{bricken2023,cunningham2023}, and circuit-level analyses have attempted to trace how such features compose into algorithms \cite{olsson2022,wang2022}. Classic vision-side counterparts include network dissection and the study of single directions important for generalisation~\cite{bau2017network,morcos2018importance}. From this standpoint, neuron selection is not merely a compression device but a probe of the organisation that training has induced---and a natural question is whether the \emph{statistical} structure of the representation, on its own, already encodes which units are functionally important.

A large body of work has attacked the efficiency question through pruning, sparsification, and regularisation. Parameters or whole neurons are ranked by criteria such as weight magnitude, local sensitivity, saliency, or estimated contribution to the loss, and the least important are removed~\cite{lecun1990second,hassibi1993optimal,han2016deep,molchanov2017variational,cheng2024survey}. Dropout regularisation has separately shown that networks remain effective when substantial fractions of hidden units are randomly suppressed during training, evidencing considerable redundancy in learned representations~\cite{srivastava2014dropout}. Importantly, this redundancy is highly non-uniform: some subnetworks or subsets of units preserve function far better than others, as emphasised by work on winning tickets, neuron importance, and the behavioural effects of compression~\cite{blalock2020state,morcos2018importance,hooker2019benchmark,frankle2019lottery}. What remains less clear is whether these unequal contributions can be inferred \emph{directly} from the internal statistical organisation of the hidden representation, rather than from supervised importance scores.

In this work we address these questions from an information-theoretic perspective, framing the selection of relevant features in a representation as a coarse-graining problem. Rather than assigning neuron importance through their immediate effect on the supervised loss, we treat the hidden layer as a configuration space generated by the network over the dataset, and define a reduced representation by retaining only a subset of hidden neurons. Such reduction can thus be interpreted as a coarsening of the space of hidden configurations, where the full representation is mapped onto a lower-dimensional description. The central question is then whether the statistical structure of these hidden configurations, by itself, is sufficient to identify informative subsets.

To quantify the effect of such reductions we employ the mapping entropy optimization workflow, or MEOW \cite{GiuliniMenichettiShellPotestio2020ModelReduction,GiuliniFiorentiniTubianaPotestioMenichetti2024EXCOGITO,aldrigo2025low,rigoli2025multiscale,guadagnin2026jctc}, an information-theoretic approach that aims at minimising the mapping entropy (ME) \cite{CG-RE-Chaimovich2010,rudzinski2011coarse,kidder2024analysis,hummerich2026}, that is a measure of the loss of distinguishability between data representations induced by a given coarse-graining. Originally developed to identify maximally informative coarse-grained representations of biomolecular systems, MEOW has recently been applied to study neural dynamics~\cite{aldrigo2025low} and the structure of parameter spaces in supervised learning problems~\cite{mele2025density}. Here, we apply MEOW to the hidden activation patterns of trained feed-forward networks. More specifically, in this setting, coarse-graining amounts to a decimation of the hidden layer, and minimising the ME allows one to identify the subset of neurons that best preserves the informational content of the representation. Neuron selection is thereby recast as an unsupervised search for reduced hidden representations that retain the relevant configurational information.

The MEOW approach provides a completely unsupervised criterion for the selection of relevant subsets of units in a hidden layer. We systematically analyse how the predictive performance of reduced networks is related to the contribution of selected neurons in preserving input information.

We start our analysis in controlled settings using synthetic datasets. In a teacher-student (TS) scenario, the structure of the hidden representation is controlled by construction and the alignment between hidden units and teacher features can be monitored explicitly~\cite{engel2001statistical,saad1995line}. This setting lets us ask precisely which neurons ME minimisation selects and how that selection evolves during training. We additionally employ a non-linear Gaussian process (NLGP) model, a synthetic but richer scenario in which hidden units spontaneously split into distinct functional classes~\cite{ingrosso2022data}; here the goal is to assess whether ME can detect this heterogeneity directly from the hidden representations. Finally, the functional significance of the selected subsets is evaluated through pruning experiments on both NLGP and MNIST trained networks, comparing the predictive performance of the reduced networks against random baselines~\cite{lecun1998gradient,Deng2012MNIST}.

\section{Methods}

The hidden layer of a feed-forward neural network is taken as the microscopic representation of the system. A coarse-grained (CG) description is then defined by selecting a subset of \(n_{cg}<K\) neurons from a hidden layer of width \(K\). Accordingly, a decimation mapping \(\mathcal{M}\) specifies which neurons are retained and therefore determines the reduced representation. In particular, given an input pattern $\mathbf{x}^\mu$, the network produces a hidden post-activation vector $\mathbf{h}^\mu = (h_1^\mu,\dots,h_K^\mu)$. Its components are binarized by taking their sign, yielding the binary hidden configuration:
\begin{equation}
\boldsymbol{\phi}^\mu = (s_1^\mu,\dots,s_K^\mu),
\qquad s_i^\mu = \mathrm{sign}(h_i^\mu) \in \{-1,+1\}.
\end{equation}

Although binarization discards potentially important magnitude information, such conservative choice allows us to reconstruct the empirical distribution without arbitrary binning and ensures that any detected structure is highly robust.

The collection of the $\boldsymbol{\phi}^\mu$ configurations over the dataset defines the ensemble of microscopic states considered in the analysis. Since the same binary configuration can occur for different input patterns, each microscopic state can be associated with an empirical probability $p(\boldsymbol{\phi})$, given by its frequency in the dataset. For each \(\boldsymbol{\phi}^\mu\), the mapping \(\mathcal{M}\) retains only selected neurons and, thus, induces a reduced configuration \(\boldsymbol{\Phi}^\mu=\mathcal{M}(\boldsymbol{\phi}^\mu)\). As a consequence, distinct microscopic configurations may be mapped onto the same reduced state and thus become indistinguishable after coarse-graining. To compare different mappings, it is therefore necessary to quantify how much of the statistical information contained in the original ensemble is lost through this reduction.

This loss is measured by the mapping entropy \cite{CG-RE-Chaimovich2010,kidder2024analysis,GiuliniFiorentiniTubianaPotestioMenichetti2024EXCOGITO,aldrigo2025low,rigoli2025multiscale,guadagnin2026jctc}. For a given mapping, the probability of a reduced configuration is obtained by summing the probabilities of all microscopic configurations mapped onto it. From this reduced description, a back-mapped distribution $\bar p_{\mathcal{M}}(\boldsymbol{\phi})$ is constructed by redistributing the probability of each reduced state uniformly among the microscopic configurations compatible with it. The ME is then defined as the Kullback--Leibler divergence between the original microscopic distribution and the distribution reconstructed from the reduced representation,
\begin{equation}
S_{\mathrm{map}}(\mathcal{M})
=
\sum_{\boldsymbol{\phi}}
p(\boldsymbol{\phi})
\log
\frac{p(\boldsymbol{\phi})}
{\bar p_{\mathcal{M}}(\boldsymbol{\phi})}.
\end{equation}

Accordingly, a smaller $S_{\mathrm{map}}$ indicates that the reduced representation preserves more of the statistical information contained in the original hidden-layer ensemble.

At fixed reduced size \(n_{cg}\), the optimal subset is identified by minimising the ME over the set \(\mathcal{M}_{n_{cg}}\) of all mappings that retain exactly \(n_{cg}\) neurons:
\begin{equation}
\mathcal{M}^\star = \arg\min_{\mathcal{M}\in\mathcal{M}_{n_{cg}}} S_{\mathrm{map}}(\mathcal{M}).
\end{equation}

The optimisation is carried out with different strategies depending on the size of the hidden layer and, correspondingly, on the number of admissible mappings. When the mapping space is sufficiently small, the minimum is determined through an exhaustive exploration of all configurations. For larger hidden layers, where such a complete enumeration becomes computationally prohibitive, the mappings minimizing the ME are searched through a stochastic simulated annealing, performed using the EXCOGITO software package implementing the MEOW protocol \cite{GiuliniFiorentiniTubianaPotestioMenichetti2024EXCOGITO, aldrigo2025low}.

\subsection{Teacher-Student}

As an initial controlled setting, we consider a one-hidden-layer network within the TS framework \citep{goldt2019dynamics,goldt2020dynamics}. Both teacher and student take inputs in $\mathbb{R}^N$ and have hidden layers of size $M$ and $K>M$, respectively. This yields a controlled over-realised TS setting, in which the student has more hidden units than the teacher, while the teacher provides a natural reference structure \citep{watkin1993statistical,engel2001statistical}. Both networks are modelled as soft committee machines with fixed second-layer weights under the standard TS normalisation, and training acts only on the student first-layer weights \citep{saad1995line,engel2001statistical}.

Two instances of this setting are considered. The first is an analytically constructed replicated configuration, in which the student hidden units are organised into groups of replicas associated with the teacher units. The second consists of trained student networks obtained by online learning on teacher-generated examples. 

The teacher first-layer weights are taken to be an orthonormal set $\{\mathbf{B}_i\}_{i=1}^{M}$, with $\mathbf{B}_i\in\mathbb{R}^N$ and $\mathbf{B}_i\cdot\mathbf{B}_j=\delta_{ij}$. Accordingly, the student has $K=nM$ hidden neurons, indexed by pairs $(i,\alpha)$, where $i\in\{1,\dots,M\}$ labels the associated teacher neuron and $\alpha\in\{1,\dots,n\}$ labels the replica within that group. Input patterns are sampled independently from the standard Gaussian distribution, $\mathbf{x}^\mu\sim\mathcal{N}(\mathbf{0},\mathbf{I}_N)$. The teacher and student outputs are defined by
\begin{align}
y_T^\mu(\mathbf{x})
&=
\frac{1}{\sqrt{M}}
\sum_{i=1}^{M}
\phi\!\left(\mathbf{x}^\mu\cdot\mathbf{B}_i\right)
=
\frac{1}{\sqrt{M}}
\sum_{i=1}^{M} t_i^\mu,
\\[1.5em]
y_S^\mu(\mathbf{x})
&=
\frac{\sqrt{M}}{K}
\sum_{i=1}^{M}\sum_{\alpha=1}^{n}
\phi\!\left(\mathbf{x}^\mu\cdot\mathbf{J}_{i,\alpha}\right)
=
\frac{\sqrt{M}}{K}
\sum_{i=1}^{M}\sum_{\alpha=1}^{n} s_{i,\alpha}^\mu,
\end{align}
where the activation function is
\begin{equation}
\phi(x)=\mathrm{erf}\!\left(\frac{x}{\sqrt{2}}\right).
\end{equation}

In the controlled setting, student first-layer weights are generated directly from the teacher weights through a tunable mismatch parameter $\eta\in[0,1]$. Specifically,
\begin{equation}
\mathbf{J}_{i,\alpha}
=
\sqrt{1-\eta^2}\,\mathbf{B}_i+\eta\,\mathbf{v}_{i,\alpha},
\end{equation}
where $\mathbf{v}_{i,\alpha}$ is a unit vector drawn uniformly in the subspace orthogonal to $\mathbf{B}_i$, so that $\mathbf{v}_{i,\alpha}\cdot\mathbf{B}_i=0$. By construction, each student vector remains normalised and has overlap
\begin{equation}
\mathbf{J}_{i,\alpha}\cdot\mathbf{B}_j
=
\sqrt{1-\eta^2}\,\delta_{ij},
\end{equation}
with the teacher weights. The parameter $\eta$ therefore controls the degree of alignment: $\eta=0$ corresponds to exact replication of the teacher representation, while $\eta=1$ corresponds to a fully orthogonal representation within the complementary subspace.

Alongside the controlled construction, trained student networks were also analysed. The student was trained by an online learning procedure on examples generated by the teacher, with only the first-layer weights updated and the second-layer weights kept fixed at their prescribed normalised values. As noted above, training naturally drives the student towards the replicated organisation that is enforced analytically in the \textit{in silico} construction. To apply the same $(i,\alpha)$ labelling to trained networks, each student neuron is assigned to the teacher class with which it has the largest overlap at the end of training. With this convention, an effective mismatch parameter is defined as
\begin{equation}
\eta^{\mathrm{eff}}_{i,\alpha}(t)
=
\sqrt{
1-
\left(
\max_j \mathbf{J}_{i,\alpha}(t)\cdot\mathbf{B}_j
\right)^2
},
\end{equation}
and its network average as
\begin{equation}
\eta_{\mathrm{eff}}(t)
=
\frac{1}{K}
\sum_{i,\alpha}
\eta^{\mathrm{eff}}_{i,\alpha}(t).
\end{equation}

The quantity $\eta_{\mathrm{eff}}(t)$ provides a time-dependent measure of the student-teacher alignment, enabling a direct comparison between trained networks and the controlled \textit{in silico} reference.

For each value of $\eta$ (or $\eta_{\mathrm{eff}}$), the ME analysis was performed over subsets of student neurons. Since the replicated construction associates each student neuron with a well-defined teacher class, the selected mappings can be characterised not only by their ME value but also by their composition. For a mapping $\mathcal{M}$, let $n_i(\mathcal{M})$ denote the number of selected student neurons associated with teacher neuron $i$, and define
\begin{equation}
p_i(\mathcal{M})=\frac{n_i(\mathcal{M})}{n_{cg}},
\qquad i=\{1,\dots,M\},
\end{equation}
with $\sum_{i=1}^{M} p_i(\mathcal{M})=1$. The balance of the mapping is then quantified by
\begin{equation}
\Delta(\mathcal{M})
=
\frac{1-\max_i p_i(\mathcal{M})}{1-1/M}.
\label{eq:delta}
\end{equation}

By construction, $\Delta=0$ when all selected neurons belong to the same class, corresponding to a maximally redundant mapping, whereas $\Delta=1$ when the selected subset is perfectly balanced across all $M$ classes. The quantity $\Delta$ was used throughout the analysis as a complementary descriptor of the structure of the selected mappings, and its definition applies without modification to both the \textit{in silico} and trained settings.

We investigated two TS configurations, both with input dimension $N=100$. In the first, the teacher and student hidden layers have sizes $M=2$ and $K=10$, respectively, and the analysis is based on $3\times 10^4$ examples; this system is small enough to allow an exhaustive exploration of the mapping space. In the second, $M=5$ and $K=25$, and the analysis is based on $10^5$ examples; in this case, the larger mapping space makes it necessary to minimise the ME by simulated annealing rather than by exhaustive search. Apart from this difference in numerical treatment, the model construction and the definitions of $\eta$, $\eta_{\mathrm{eff}}$, and $\Delta$ are identical in the two settings.

\subsection{Non-Linear Gaussian Process}

As a second model, we considered a synthetic binary classification task based on a NLGP \citep{ingrosso2022data}. Input patterns are $N$-dimensional vectors, where each component $x_i$, with $i=1,\dots,N$, plays the role of a pixel intensity. Two classes are generated, corresponding to Gaussian fields with correlation lengths $\xi^+$ and $\xi^-$, respectively. The patterns belonging to each class are generated as
\begin{equation}
    \mathbf{x}^{\xi^\pm}
    =
    \frac{\psi(\gamma\,\mathbf{z}^{\xi^\pm})}{Z(\gamma)},
    \label{eq:nlgp_data}
\end{equation}
where $\mathbf{z}^{\xi^\pm}\in\mathbb{R}^N$ is a zero-mean Gaussian vector with covariance
\begin{equation}
    C^{\xi^\pm}_{ij}
    =
    \langle z^{\xi^\pm}_i z^{\xi^\pm}_j\rangle
    =
    \exp\!\left(-\frac{|i-j|^2}{(\xi^{\pm})^2}\right),
    \label{eq:nlgp_cov}
\end{equation}
with periodic boundary conditions imposed on the input coordinate index. The two classes are therefore distinguished by their respective correlation lengths $\xi^+$ and $\xi^-$. The non-linear function $\psi(z)=\mathrm{erf}(z/\sqrt{2})$ introduces non-Gaussian statistics controlled by the parameter $\gamma$, while the normalisation factor $Z(\gamma)$ is chosen such that $\mathrm{Var}(\mathbf{x}^{\xi^\pm})=1$. Throughout the analysis, the values $N=50$, $\xi^+=7.5$, $\xi^-=2$, and $\gamma=15$ were employed. Both the training and test sets contain a total of $P=\alpha N$ patterns, with $\alpha=300$, equally divided between the two classes. The labels assigned to the two classes are $y=+1$ and $y=-1$, respectively to configurations with correlation length $\xi^+$ and $\xi^-$.

The network is a one-hidden-layer perceptron with $N$ input units, $K=30$ hidden neurons, and a single linear output unit. The hidden-layer activation function is
\begin{equation}
    \phi(x)=\mathrm{erf}\!\left(\frac{x}{\sqrt{2}}\right),
\end{equation}
and the output weight vector is fixed to $w^{\mathrm{out}}_i=1/\sqrt{K}$ for all $i$, with zero output bias; only the first-layer weights and biases are trained. Training was performed by stochastic gradient descent on the mean squared error loss with $L_2$ regularisation coefficient $0.1$, learning rate $0.1$, and batch size $300$. Networks were trained for up to $3\times 10^4$ epochs.

For the analysis of trained networks, hidden neurons were partitioned into two empirical groups according to the structure of their incoming weight vectors. Neurons with weight vectors concentrated on a restricted subset of input coordinates were classified as localised, whereas neurons with extended alternating-sign profiles were classified as oscillatory \citep{ingrosso2022data}. The two groups were quantitatively identified by means of the inverse participation ratio (IPR), defined for each hidden neuron $i$ as
\begin{equation}
    \mathrm{IPR}(\mathbf{w}_i)
    =
    \frac{\sum_{j=1}^{N} w_{ij}^4}
         {\left(\sum_{j=1}^{N} w_{ij}^2\right)^2},
    \label{eq:ipr}
\end{equation}
which ranges between $1/N$, attained when all weights are equal in magnitude, and $1$, attained when only a single weight is non-zero. Localised neurons therefore exhibit systematically larger $\mathrm{IPR}$ values than oscillatory ones. The two groups also develop distinct bias values during training, with oscillatory neurons associated with positive bias and localised neurons with negative one, thus providing a complementary indicator for their identification and tracking.

The ME analysis was performed at multiple checkpoints during training. At each epoch, the hidden representations were computed by passing all training patterns through the network and binarising the continuous activations via the sign function. The ME minimisation was then carried out independently at each checkpoint and for each value of the number of neurons retained $n_{\mathrm{cg}}$, using multiple independent stochastic minimisation runs. Additionally, the composition of the selected mappings can be also characterised in terms of the fraction of localised neurons $f_{\mathrm{loc}}(\mathcal{M})$ among the $n_{\mathrm{cg}}$ retained units,
\begin{equation}
    f_{\mathrm{loc}}(\mathcal{M})
    =
    \frac{n_{\mathrm{loc}}(\mathcal{M})}{n_{cg}}.
\end{equation}

\subsection{Wang-Landau Sampling of the Mapping Space}

In addition to the direct minimisation of the ME, the structure of the mapping space of the NLGP system was characterised via Wang--Landau (WL) sampling in the $1/t$ formulation \citep{wang2001efficient, belardinelli2007wang,  mele2025density}. While ME minimisation identifies mappings with minimal information loss, it does not provide information on the global organisation of the mapping space. WL sampling complements this analysis by estimating the density of states (DoS), i.e., the number of configurations associated with a given value of an effective energy. The DoS is computed by adaptively biasing the Monte Carlo sampling to promote a broad exploration of the effective energy range while iteratively refining the current DoS estimate. The magnitude of these updates is controlled by a refinement parameter, which is progressively reduced during the simulation according to the \(1/t\) scheme. In the present application, a configuration corresponds to a mapping $\mathcal{M}$, and the ME, $S_{\mathrm{map}}(\mathcal{M})$, is taken as its effective energy. Sampling was terminated when the refinement parameter reached $10^{-5}$.

At fixed retained neuron size $n_{\mathrm{cg}}$ and fixed number of localised neurons $n_{\mathrm{loc}}$, the mapping space is defined as the set of all subsets of $n_{\mathrm{cg}}$ hidden neurons containing exactly $n_{\mathrm{loc}}$ localised units or, analogously, $n_{\mathrm{cg}} - n_{\mathrm{loc}}$ oscillatory units.

The WL analysis was performed at fixed $n_{\mathrm{cg}} = 15$. This choice is particularly relevant for the present system, as the hidden layer consists of $30$ neurons, evenly divided into $15$ localised and $15$ oscillatory units, making $n_{\mathrm{cg}} = 15$ a natural CG size for the analysis. For each admissible value of $n_{\mathrm{loc}}$, an independent WL simulation was carried out on the corresponding constrained mapping space. Trial moves were defined by exchanging one (or more) retained neuron(s) with one (or more) discarded neuron(s) belonging to the same class, thereby preserving both $n_{\mathrm{cg}}$ and $n_{\mathrm{loc}}$ throughout the sampling dynamics. Each simulation was continued until the refinement parameter reached $10^{-5}$.

The quantity of interest is the joint DoS:
\begin{equation}
g\!\left(S_{\mathrm{map}}, n_{\mathrm{loc}}\right),
\end{equation}
or, equivalently, its representation in terms of the fraction of localised neurons $f_{\mathrm{loc}} = \frac{n_{\mathrm{loc}}}{n_{\mathrm{cg}}}$. The joint DoS was reconstructed by combining the independently sampled spectra obtained at fixed $n_{\mathrm{loc}}$. This reconstruction is possible because each simulation explores the full range of effective energies accessible under the corresponding compositional constraint and because, for fixed $n_{\mathrm{cg}}$ and $n_{\mathrm{loc}}$, the total number of admissible mappings is known exactly. These combinatorial counts provide the appropriate normalisation for each restricted DoS, thereby allowing the different contributions to be assembled into the full two-dimensional DoS. The same analysis was performed at different training epochs in order to characterise the evolution of the mapping space structure during learning.

\subsection{Performance Evaluation of Reduced Networks}

To test whether the mappings selected by ME minimisation also identify effective reduced representations, we construct pruned networks by retaining only the hidden neurons selected by the mapping. For each \(n_{\mathrm{cg}}\) value, the performance of the selected subset is compared with that obtained from random subsets of the same size. In the NLGP case, pruning is applied to the hidden layer of the trained classifier. After selecting the retained neurons, the output bias is re-optimised while keeping the hidden-layer parameters fixed. 

As a second classification benchmark, we consider a feed-forward neural network with a single hidden layer of \(K=30\) neurons and \(\mathrm{erf}\) activation function. The task is binary classification on the MNIST dataset \citep{lecun1998gradient, Deng2012MNIST}, restricted to digits \(1\) and \(7\). To make the problem translationally invariant and more challenging, the dataset is augmented by random translations of the input images. The network is trained by stochastic gradient descent, with only the first-layer parameters optimised. Reduced networks are then obtained by pruning the hidden layer according to the selected mapping, and their test accuracy is compared with that obtained from random subsets of the same size.

\section{Results}

\subsection{Structure of the Mappings Selected by Mapping Entropy Minimisation}

In this section, we analyse the properties of the subsets selected by ME minimisation in controlled settings where the organisation of the hidden representation is already understood. To this end, we consider two complementary scenarios: the overparameterised TS regression model, where hidden units can be interpreted through their alignment with the teacher's units, and the NLGP classification problem, where neurons organise into distinct functional classes. In both cases, this well-defined structure, known \emph{a priori}, makes it possible to interpret the subsets selected by ME.

\subsubsection{Teacher-Student system}

\begin{figure*}[t]
    \centering
    \includegraphics[width=\linewidth]{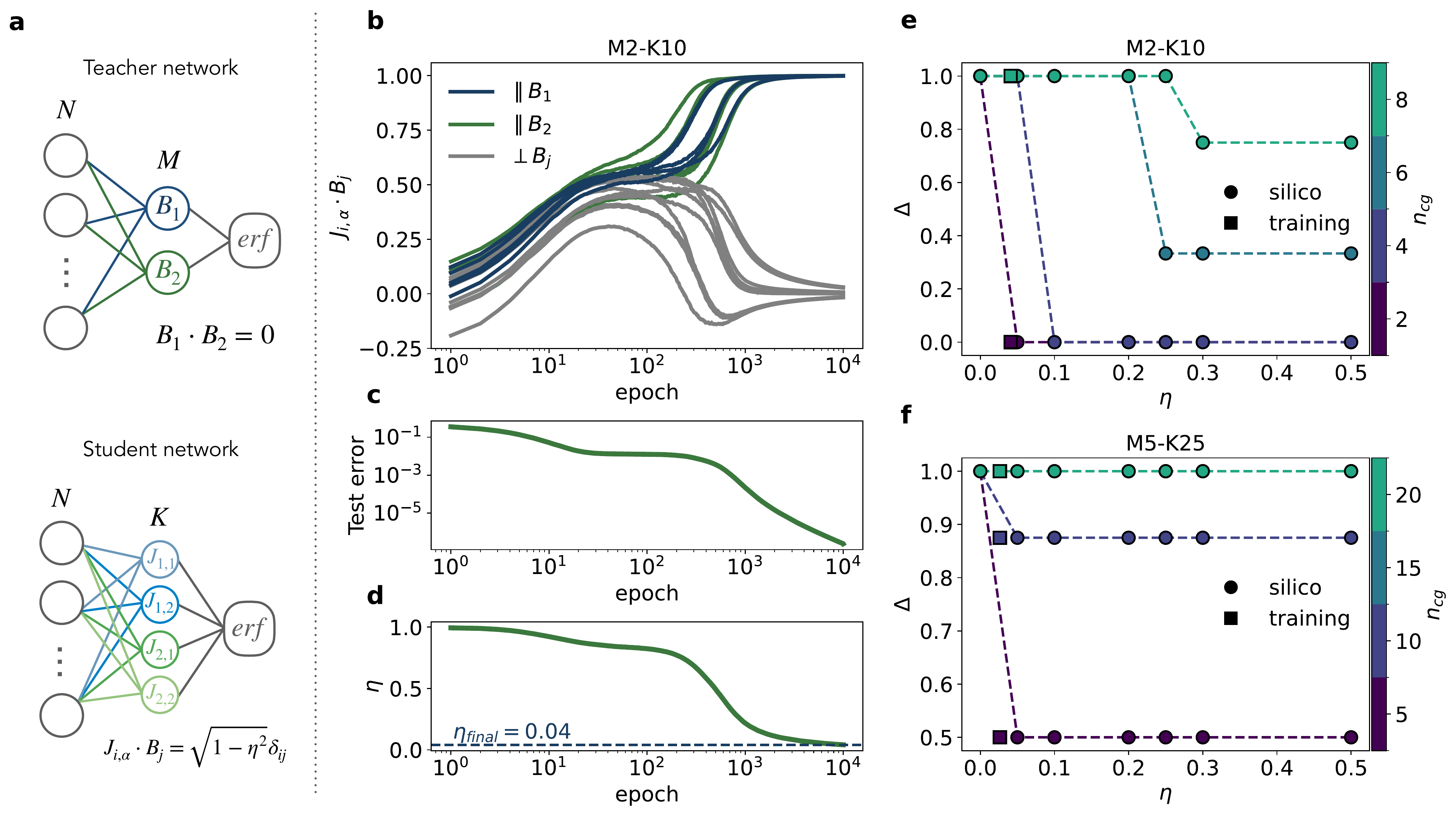}
    \caption{ TS system and ME analysis of hidden-layer representations. (\textbf{a}) Schematic representation of the teacher and student networks. The teacher has \(M\) hidden units with orthogonal first-layer weights \(\mathbf{B}_i\), while the student has \(K>M\) hidden units labelled by \((i,\alpha)\). In the controlled construction, the student weights satisfy \(\mathbf{J}_{i,\alpha}\!\cdot\!\mathbf{B}_j=\sqrt{1-\eta^2}\,\delta_{ij}\). (\textbf{b}) Training dynamics of the overlaps \(\mathbf{J}_{i,\alpha}\!\cdot\!\mathbf{B}_j\) in the M2-K10 case, shown as a function of the training epoch. Blue and green curves denote the overlaps with \(\mathbf{B}_1\) and \(\mathbf{B}_2\), respectively, while grey curves denote the components orthogonal to the teacher directions. (\textbf{c}) Generalization error as a function of the training epoch for the same system. (\textbf{d}) Effective mismatch parameter \(\eta_{\mathrm{eff}}\) as a function of the training epoch. The dashed horizontal line indicates the final value \(\eta_{\mathrm{final}}=0.04\). (\textbf{e}, \textbf{f}) Balance \(\Delta\) of the ME-minimising mapping as a function of the mismatch parameter \(\eta\), for the M2-K10 and M5-K25 systems, respectively. Different colours identify different values of the retained CG dimensionality \(n_{\mathrm{cg}}\), as indicated by the colour bar. Circular markers denote the controlled \textit{in silico} construction, and square markers denote the trained networks, plotted at the corresponding effective mismatch value \(\eta_{\mathrm{eff}}\). Note the different scale of the $\Delta$ axis in the two plots.}
    \label{fig:TS}
\end{figure*}

We begin with the controlled TS setting, where the structure of the student hidden layer is fixed by construction and therefore provides a direct reference for interpreting the subsets selected by ME minimisation. The relevant observable is the balance index \(\Delta\), defined in Eq. \ref{eq:delta}, whose values are shown in \autoref{fig:TS}e,f as a function of the mismatch parameter \(\eta\) for different numbers of retained neurons \(n_{\mathrm{cg}}\). By construction, \(\Delta=1\) identifies a mapping that distributes the retained neurons as evenly as possible across teacher groups, whereas smaller values correspond to progressively more concentrated selections. \(\Delta\) values close to $1$ are therefore desirable, because they indicate that the selected subset preserves the full teacher structure rather than over-representing only a restricted part of it.

We start from the exact replicated limit, \(\eta=0\), which provides the natural reference case for the discussion. Here, the ME-minimising mapping is fully balanced (\(\Delta=1\)) at the smallest number of neurons compatible with the teacher structure: \(n_{\mathrm{cg}}=2\) for the M2-K10 system and \(n_{\mathrm{cg}}=5\) for the M5-K25 system (\autoref{fig:TS}.e,f). Thus, when the student exactly replicates the teacher representation, the entropy minimum coincides with the smallest subset that still covers all teacher modes. This is a non-trivial result, because the ME is computed solely from the statistics of the hidden configurations and has no explicit access to the teacher labels; nevertheless, in the replicated limit it selects one representative degree of freedom for each teacher mode.

As soon as the mismatch becomes finite, deviations from this correspondence begin to emerge. In both systems, the curves in \autoref{fig:TS}.e,f show that the smallest number of neurons for which the selected mapping is balanced increases monotonically with \(\eta\). In the M2-K10 case, the mapping selected at the minimal size \(n_{\mathrm{cg}}=2\) is balanced only at \(\eta=0\), while already at \(\eta \simeq 0.1\) it has collapsed to \(\Delta=0\). By contrast, the first \(n_{\mathrm{cg}}\) value that remains balanced over an extended mismatch range is \(n_{\mathrm{cg}}=4\), which keeps \(\Delta=1\) up to \(\eta \approx 0.2\). A similar trend is visible in the M5-K25 case, but on a broader scale: the minimal balanced mapping occurs at \(n_{\mathrm{cg}}=5\) only at \(\eta=0\), while for finite mismatch the balanced solution is shifted to substantially larger numbers of retained neurons. For instance, curves with \(n_{\mathrm{cg}}\approx 10\) remain at \(\Delta \simeq 0.875\), showing that they still over-represent some teacher groups, whereas the first fully balanced solution appears only for \(n_{\mathrm{cg}}=15\).

This systematic shift has a clear interpretation. The ME does not probe functional equivalence with the teacher, but the statistical distinguishability of the hidden units. At \(\eta=0\), different replicas associated with the same teacher unit are statistically indistinguishable, so one representative per group is sufficient to fully reproduce the whole network function. At finite \(\eta\), each student neuron acquires an additional orthogonal component and replicas are no longer equivalent at the level of hidden-layer statistics. As the variability of the hidden representation increases, the overall minimum of this profile shifts towards larger \(n_{\mathrm{cg}}\) values, because more neurons must be retained to resolve the additional variability. In this sense, the increase of the optimal \(n_{\mathrm{cg}}\) is not simply a failure to recover the teacher-minimal representation; rather, it quantifies how far the hidden representation is from the exact replicated limit.

It is then natural to ask whether the same scenario survives in trained networks, where the variability is not imposed homogeneously by construction but generated by the learning dynamics. Panels \autoref{fig:TS}.b--d show that training indeed drives the student towards the teacher representation. In the M2-K10 example, the overlaps with the two teacher directions increase from values close to zero at the beginning of training to values close to one at late times, while the orthogonal components are progressively suppressed (\autoref{fig:TS}.b). Over the same interval, the generalisation error decreases by several orders of magnitude (\autoref{fig:TS}.c), and the effective mismatch \(\eta_{\mathrm{eff}}\) drops from approximately \(1\) to a final value \(\eta_{\mathrm{final}}\simeq 0.04\) (\autoref{fig:TS}.d). The trained student therefore approaches, but does not exactly reach, the replicated limit.

The ME analysis of the trained networks is fully consistent with the controlled construction. In \autoref{fig:TS}.e,f, the square markers corresponding to the trained networks fall on the same branches defined by the \textit{in silico} curves when plotted at the corresponding value of \(\eta_{\mathrm{eff}}\). Quantitatively, the trained networks no longer recover a balanced mapping at the strictly minimal size \(n_{\mathrm{cg}}=M\): for M2-K10 the smallest balanced mapping is found at \(n_{\mathrm{cg}}=4\), and for M5-K25 at \(n_{\mathrm{cg}}=15\). These are precisely the same \(n_{\mathrm{cg}}\) values selected by the controlled curves at mismatch values of order \(\eta \simeq 0.04\), showing that the trained networks behave as weakly mismatched replicas of the teacher representation.

Taken together, these results identify a clear selection principle in the TS setting. In the exact replicated limit, ME minimisation recovers the minimal teacher-consistent hidden representation. As soon as replicas are no longer statistically equivalent, the \(n_{\mathrm{cg}}\) value for which  $\Delta=1$ increases, and the amount by which it increases provides a direct measure of the residual variability present in the hidden layer. The agreement between the analytically controlled construction and the trained networks shows that this interpretation remains valid beyond the idealised setting and continues to hold in the presence of heterogeneous, learning-induced correlations.

\subsubsection{NLGP System}

\begin{figure*}[t]
    \centering
    \includegraphics[width=\linewidth]{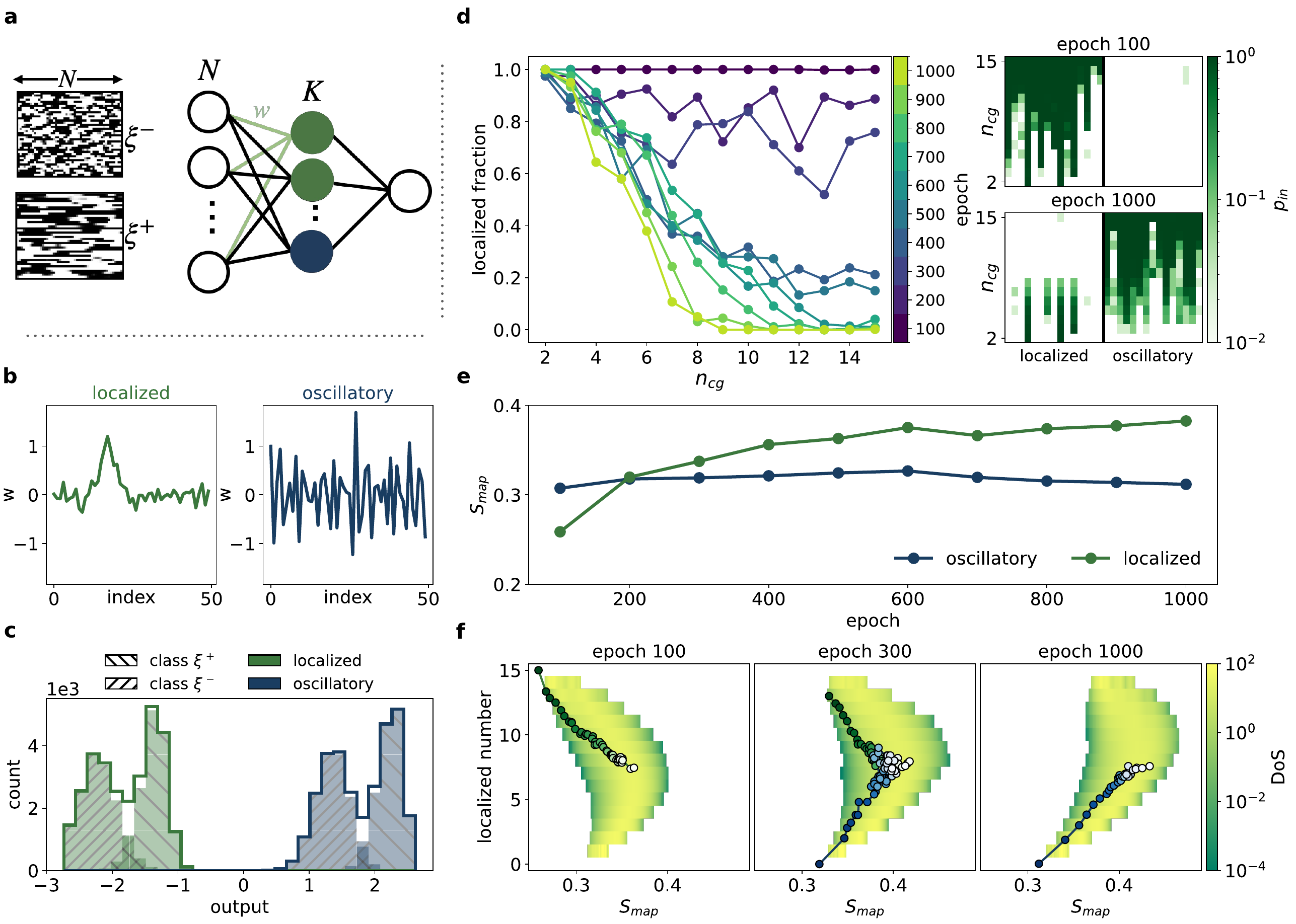}
    \caption{NLGP system and ME analysis of hidden-layer representations. 
(\textbf{a}) Schematic of the classification task and network architecture. Input patterns from the two classes, characterised by different correlation lengths, are shown alongside the one-hidden-layer network with \(K=30\) neurons. 
(\textbf{b}) Representative examples of hidden neurons after training: a localised neuron (left), with weights concentrated on a subset of input coordinates, and an oscillatory neuron (right), with extended alternating-sign structure. 
(\textbf{c}) Distribution of the network output evaluated using only localised neurons (green) or only oscillatory neurons (blue), compared across the two classes (hatched histograms). 
(\textbf{d}) Left: average fraction of localised neurons in the ME-minimising mapping as a function of the number of retained neurons \(n_{\mathrm{cg}}\), for different training epochs (colour-coded). Right: selection probability \(p_{\mathrm{in}}\) of each neuron, shown as a function of \(n_{\mathrm{cg}}\), for two representative epochs (100 and 1000). Neurons are ordered according to their bias value. 
(\textbf{e}) ME \(S_{\mathrm{map}}\) evaluated for mappings composed exclusively of localised neurons (green) or oscillatory neurons (blue), as a function of training epoch. 
(\textbf{f}) DoS of the mapping space as a function of \(S_{\mathrm{map}}\) and the number of localised neurons, for different training epochs. Overlaid points indicate average optimisation trajectories.
}
    \label{fig:NLGP}
\end{figure*}

A richer and less constrained setting is provided by the NLGP classification problem, where the hidden representation is structured, but not through explicit replication. The full set of results is shown in \autoref{fig:NLGP}. As already known from previous studies \citep{ingrosso2022data}, after training the hidden neurons separate into two classes, illustrated in \autoref{fig:NLGP}.b: localised neurons, whose weights are concentrated on a restricted subset of input coordinates, and oscillatory neurons, whose weights display an extended alternating-sign profile. Here this prior structural information is used as a reference frame to analyse the mappings selected by the ME minimisation.

Before turning to the ME analysis, it is important to establish that the two neuron classes correspond to distinct yet individually informative representations of the classification task. This is shown in \autoref{fig:NLGP}.c, where the network output is evaluated after retaining only one class of hidden neurons at a time. In both cases, the outputs associated with the two input classes remain clearly separated, showing that either population alone is sufficient to preserve discriminative information. The difference lies in the form of the representation: localised neurons and oscillatory neurons induce distinct output distributions, reflecting two qualitatively different internal encodings of the same classification rule. \autoref{fig:NLGP}.c therefore shows that the two populations should not be interpreted as playing complementary roles that become meaningful only when combined; rather, they define alternative representational modes, each of which can by itself support the separation of the input classes.

The ME analysis shows that the optimal mappings do not combine these two representations arbitrarily. In \autoref{fig:NLGP}.d (left subpanel) we report the average fraction of localised neurons among the optimal mappings for different training epochs. Early in the training phase the neurons selected in the mapping are almost invariably localised irrespective of the subset size \(n_{\mathrm{cg}}\). This picture is confirmed by the selection probability \(p_{\mathrm{in}}\) that a given neuron is included in an optimal mapping, reported in the right subpanel of \autoref{fig:NLGP}.d: at epoch \(100\), in fact, the optimisation repeatedly selects the same band of localised neurons with \(p_{\mathrm{in}}\approx 1\), whereas oscillatory neurons are almost never included.

As training proceeds, this preference changes in a strongly size-dependent way. Around epoch \(300\), the average localised fraction starts to decrease for intermediate and large \(n_{\mathrm{cg}}\) values, signalling the appearance of a competing oscillatory solution. By late training, the transition is sharp: at epoch \(1000\), the optimal mappings are still almost fully localised for very small subsets, but the localised fraction drops below \(1/2\) already around \(n_{\mathrm{cg}}\simeq 6\!-\!7\), and becomes essentially zero for \(n_{\mathrm{cg}}\gtrsim 10\). The selection probability map at epoch \(1000\) shows that this crossover does not arise from a gradual mixing of the two populations. Rather, for small \(n_{\mathrm{cg}}\) the selected neurons come almost exclusively from the localised group, whereas for larger \(n_{\mathrm{cg}}\) the selected mappings are almost entirely oscillatory.

This interpretation is corroborated by the direct comparison shown in \autoref{fig:NLGP}.e at \(n_{\mathrm{cg}}=15\), where the retained subset coincides with the full set of oscillatory neurons or, alternatively, with the full set of localised neurons. At epoch \(100\), purely localised mappings have a substantially lower ME than purely oscillatory ones, with \(S_{\mathrm{map}}\approx 0.26\) for the localised case against \(S_{\mathrm{map}}\approx 0.31\) for the oscillatory one. Around epoch \(200\), the two values become comparable, \(S_{\mathrm{map}}\simeq 0.32\), marking the point at which the two representational modes compete on equal footing. At later epochs the ordering is reversed: the value of the ME of selections containing exclusively oscillatory neurons remains approximately flat in the interval \(0.31\!-\!0.33\), whereas the one including only localised neurons increases steadily up to \(\approx 0.38\) by epoch \(1000\). The growing gap quantitatively explains why, at late times, oscillatory mappings dominate the optimal set at large \(n_{\mathrm{cg}}\).

The WL reconstruction of the full mapping space provides a more global view of this transition. In \autoref{fig:NLGP}.f, the DoS is shown as a function of \(S_{\mathrm{map}}\) and the number \(n_{\mathrm{loc}}\) of localised neurons retained in mappings of fixed size \(n_{\mathrm{cg}}=15\). A first important point is that the ME minima are generally not found in mixed mappings containing substantial contributions from both functional classes; rather, they lie close to the edges of the space, where the retained subset is dominated by one class or the other. At epoch \(100\), the low-$S_{\mathrm{map}}$ basin is found at large \(n_{\mathrm{loc}}\), at the fully localised edge, and the optimisation trajectories converge to this region. At epoch \(300\), the landscape becomes effectively bimodal: two competing basins are visible, and the optimisation trajectories split between them. By epoch \(1000\), the low-$S_{\mathrm{map}}$ basin has moved to \(n_{\mathrm{loc}}=0\), \ie to purely oscillatory mappings, while the localised side remains only as a higher-$S_{\mathrm{map}}$ sector of the mapping space. Therefore, the change observed in \autoref{fig:NLGP}.d is not a finite sampling artefact of the annealing dynamics, but it reflects a true restructuring of the ME landscape.

Overall, the NLGP results show that ME minimisation does not favour sparse or mixed subsets. Instead, it selects coherent mappings associated with one of the two pre-existing representational classes, and the preferred class depends both on the stage of training and on the number of neurons retained. At early epochs, localised neurons provide the most informative reduced description. During training, an oscillatory representation progressively becomes competitive and eventually dominates for sufficiently large \(n_{\mathrm{cg}}\). The ME therefore acts as a probe of the internal organisation of the hidden layer, resolving which representational mode provides the statistically more significant CG description.

\subsection{Performance of the Reduced Network}
Having characterised the structure of the mappings selected by ME minimisation in controlled settings, we now turn to their functional validation. The goal is to determine whether these subsets, beyond exhibiting a clear and non-trivial organisation in the benchmark cases discussed above, also preserve the predictive performance of the original network after pruning. Since the MEOW criterion relies solely on the statistics of the hidden activation patterns, with no direct access to the task or to the output labels, this constitutes a stringent test of its effectiveness as an unsupervised pruning criterion.

\begin{figure}
  \centering
  \includegraphics[width=\linewidth]{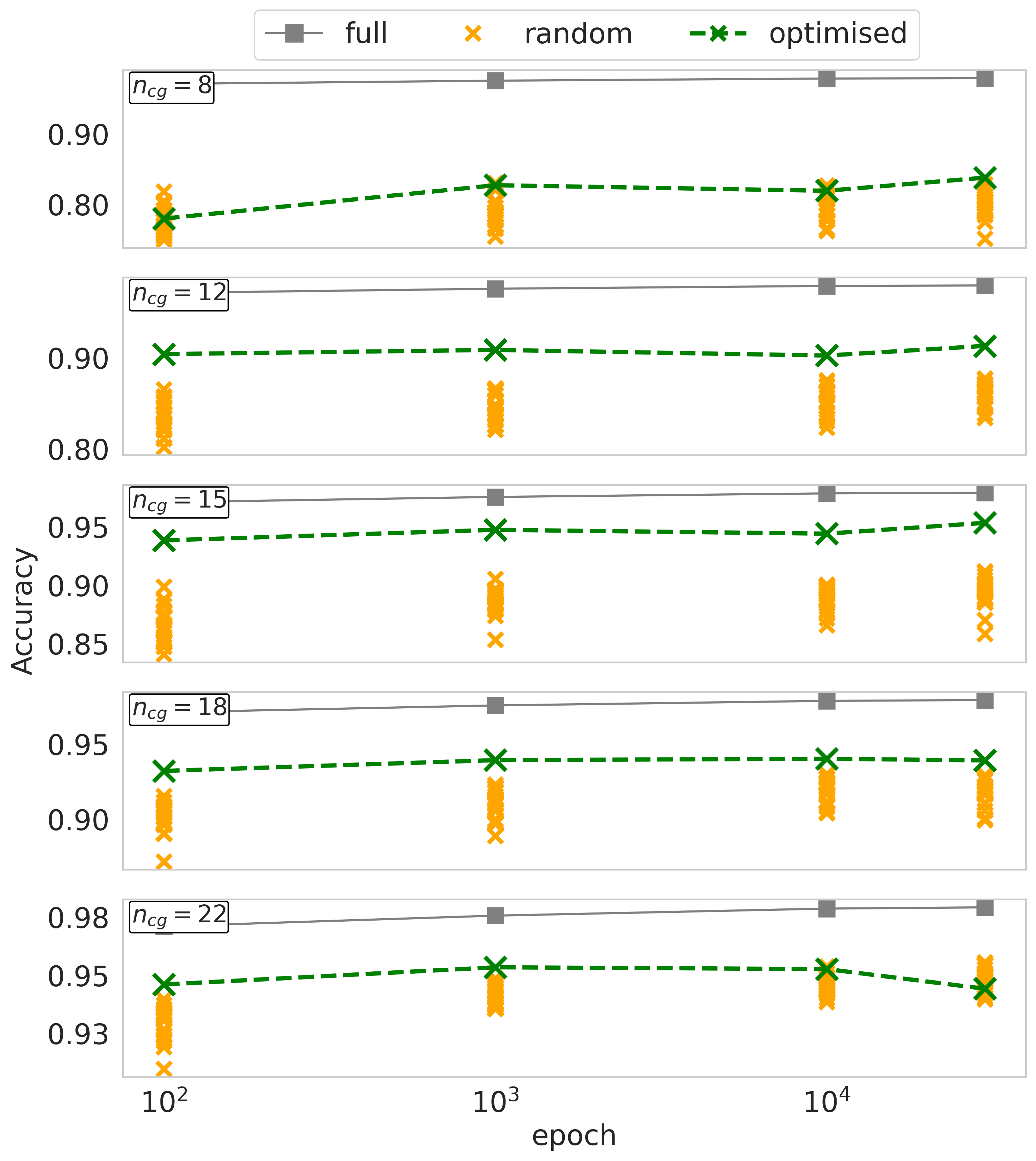}
  \caption{NLGP reduced-network performance at different training epochs. Classification accuracy as a function of the number of retained neurons \(n_{\mathrm{cg}}\). For each \(n_{\mathrm{cg}}\), the green markers denote the accuracy of the reduced network obtained from the ME-selected subset, the orange markers the accuracies obtained from random subsets of the same size, and the dashed grey line the accuracy of the full network. After pruning, the output bias is re-optimised while keeping the hidden-layer parameters fixed. Note that the $y$ axes do not start from zero.}
  \label{fig:nlgp_acc}
\end{figure}

We begin with the NLGP classifier, where the performance of reduced networks can be monitored at different stages of training and for several values of retained neurons \(n_{\mathrm{cg}}\) (\autoref{fig:nlgp_acc}). For each \(n_{\mathrm{cg}}\) value, the accuracy obtained from the MEOW-selected subset is compared with that of random subsets of the same cardinality, as well as with the accuracy of the full network. After pruning, the output bias is re-optimised while keeping the hidden-layer parameters fixed. This is a minimal readjustment, introduced to compensate for the shift of the output distribution induced by the selected hidden subset. Its use is motivated looking at \autoref{fig:NLGP}.c: localised and oscillatory populations define distinct, yet individually viable, representations of the classification rule, but they generally induce different offsets in the output distribution. As a consequence, pruning changes the effective decision threshold of the classifier. If the zero threshold used for the full network were kept unchanged, all reduced selections would yield an accuracy close to \(0.5\), with one of the two classes being almost entirely misclassified. Re-optimising the output bias therefore removes this trivial source of performance loss and allows one to compare different subsets on the basis of the information retained in the hidden representation.

From an inspection of \autoref{fig:nlgp_acc}, a clear trend emerges. For all epochs considered, the ME-selected subsets typically achieve an accuracy above the bulk of the random distribution, showing that the mappings identified through the activation statistics also retain a comparatively informative representation for the task. This advantage is most visible in the strongly compressed regime. For instance, at \(n_{\mathrm{cg}}=8\) and \(n_{\mathrm{cg}}=12\), the optimised subset lies systematically above the typical random choice at all epochs. The same behaviour persists at \(n_{\mathrm{cg}}=15\) and \(18\), although the separation becomes progressively smaller as the number of retained neurons increases. By contrast, at \(n_{\mathrm{cg}}=22\) the MEOW selection and the random choice of retained neurons provide nearly indistinguishable outcomes, as expected when most of the hidden layer is already retained. Overall, the NLGP results indicate that ME minimisation provides a non-trivial and functionally meaningful guide for pruning, especially when the compression is so strong (i.e. when up to half of the hidden layer neurons are culled) that the specific choice of subset matters.

\begin{figure}
  \centering
  \includegraphics[width=\linewidth]{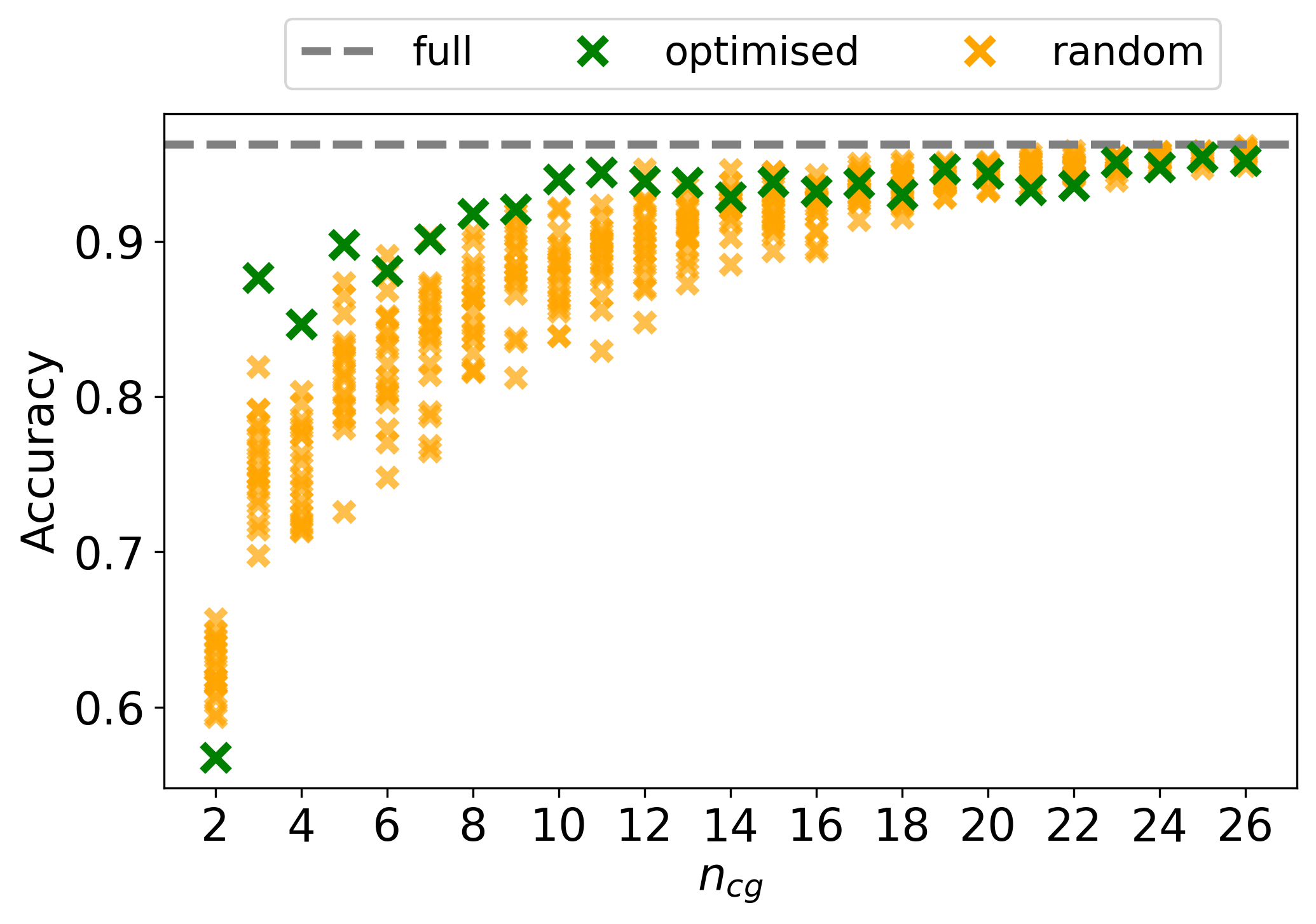}
  \caption{Reduced-network performance for the MNIST binary classification task (\(1\) vs \(7\)) with data augmentation by random translations. Classification accuracy as a function of the number of retained neurons \(n_{\mathrm{cg}}\). For each \(n_{\mathrm{cg}}\), the green markers denote the accuracy of the reduced network obtained from the ME-selected subset, the orange markers the accuracies obtained from random subsets of the same size, and the dashed grey line the accuracy of the full network. Note that the $y$ axes do not start from zero.}
  \label{fig:mnist_acc}
\end{figure}

The same analysis can then be extended to a more realistic classification problem, namely the binary MNIST task with data augmentation by random translations, using a network with $K=30$ hidden-neurons. In this case, shown in \autoref{fig:mnist_acc}, we report the reduced-network performance only at the end of training, since the behaviour is qualitatively stable across epochs. Also here the accuracy obtained from the MEOW-selected subset is compared with the distribution generated by random subsets of equal size. The overall picture is consistent with the NLGP benchmark, but with a clearer dependence on the number of retained neurons. Excluding the extreme case \(n_{\mathrm{cg}}=2\), where all reduced networks perform poorly, the ME-selected mappings systematically lie in the upper part of the random distribution for small and intermediate \(n_{\mathrm{cg}}\) values. In particular, for \(3 \lesssim n_{\mathrm{cg}} \lesssim 12\), the selected subset typically yields an accuracy well above the bulk of the random configurations and often close to their upper tail. This indicates that, in the strongly pruned regime, the ME criterion is able to identify hidden subsets that preserve most effectively the task-relevant information. For larger \(n_{\mathrm{cg}}\) values, however, the distinction progressively weakens: the random and optimised subsets become compatible within the observed spread. This is consistent with the fact that, when the reduction is mild, the space of admissible subsets becomes much less heterogeneous and the benefit of a structured selection is correspondingly reduced.

Taken together, these results show that the mappings selected by ME minimisation are not only structurally meaningful, but also functionally relevant. The selection criterion does not identify the optimal reduced network in an absolute sense, nor should this be expected, since the approach is blind to the specific task and relies only on the hidden activation statistics; nevertheless, in a broad range of regimes it provides an effective proxy for pruning. Its advantage is most evident when the compression is sufficiently strong that the choice of the retained subset is non-trivial, while it naturally diminishes as \(n_{\mathrm{cg}}\) approaches the size of the full hidden layer. Because the MEOW protocol is strictly unsupervised, we omit comparisons with supervised pruning measures (e.g., weight magnitude, saliency, or gradient scores) that exploit information ME deliberately ignores. This isolates the capacity of activation statistics alone to identify critical units, leaving benchmarks against supervised criteria for future work.

\section{Conclusions}

In this work, we have investigated the space of coarse-grainings of neural networks through the lens of the MEOW approach, with the aim of identifying reduced representations that are both structurally meaningful and functionally effective. Our information-theoretic framework yields a purely unsupervised method to select informative subsets of neurons in a network, highlight functional heterogeneity within hidden layers and identify the evolution of learned representations throughout training. Minimizing the mapping entropy thus presents itself as a simple strategy to identify relevant features of the internal representation.

The proposed strategy was put at test in multiple and diverse contexts. In controlled TS settings, the selected neurons are strongly aligned with the informative directions defined by the teacher. In the NLGP classification task, where the hidden layer spontaneously organises into two functionally distinct populations, the method does not select generic high-variance units: it selects coherent mappings drawn from a single representational class, and the preferred class changes over the course of training.

Crucially, reduced networks obtained by retaining the neurons selected through ME minimisation consistently outperform those constructed from random subsets of the same size. This establishes a concrete link between information-theoretic optimality and predictive performance, showing that the preservation of configurational distinguishability translates into better-preserved predictive performance relative to random pruning: at equal compression, ME-selected subsets degrade markedly less than random ones. These findings suggest that the ME captures structural information that is intrinsic to the data representation, rather than being tied to a specific task or loss function. In this sense, the MEOW strategy represents a principled pruning criterion that is complementary to standard approaches based on weight magnitude, sensitivity analysis, or gradient-based importance measures \citep{lecun1990second,molchanov2017variational}. Unlike these methods, which rely on supervised signals, the MEOW approach operates directly on the statistical structure of the hidden configurations, making it applicable in a broader range of settings.

Several concrete directions stem from this work, which we detail below.

\paragraph{Scaling and the lazy-to-rich transition.}
An interesting, relevant avenue would be the development of a simple representation-level diagnostic that is sensitive to the crossover from lazy to rich behaviour. In fact, a natural next step for our work is to scale the approach to substantially larger and wider networks and to use it as a diagnostic for the crossover between the feature-learning and lazy (kernel/NTK) regimes~\cite{jacot2018,chizat2019,geiger2020}. Because the ME is sensitive to whether hidden units are statistically distinguishable or interchangeable, we expect that the ME landscape, and in particular the balance and composition of ME-optimal subsets, should behave qualitatively differently in the two regimes: lazy-regime representations, inherited from initialisation and only weakly reorganised, should exhibit a flatter, more degenerate selection landscape than feature-learning ones. Systematically varying width, initialisation scale, and learning rate would allow the ME to be assessed as a representation-level order parameter for locating this transition.

\paragraph{Iterative pruning for deep networks.}
The present study prunes a single hidden layer in one shot. A promising extension of our protocol is an \emph{iterative} ME-based pruning scheme for deep architectures, in which units are removed in several rounds, each time re-estimating the hidden configuration statistics and re-minimising the ME on the surviving representation, in the spirit of iterative magnitude pruning and the lottery-ticket procedure \cite{blalock2020state,frankle2019lottery}. Interleaving decimation with brief fine-tuning between rounds could compensate for the shift of the internal representation and allow much stronger compression than single-shot selection.

\paragraph{Other layers and architectures.}
Our framework applies to any layer that produces a well-defined configuration ensemble, so it can be carried beyond single fully-connected layers to convolutional feature maps and, most interestingly, to the internal representations of attention-based models and transformers. On the methodological side, such extensions call for more general forms of coarse-graining, going beyond the simple selection of individual neurons employed in this work. Applying ME-based coarse-graining to the residual stream, attention heads, or MLP neurons of transformers would connect this line of work to mechanistic interpretability, where superposition and polysemanticity make principled, unsupervised notions of unit importance particularly valuable \cite{elhage2022,bricken2023}; the ME minimisation could offer a statistics-driven criterion for head or feature selection that is complementary to dictionary-learning approaches.

\paragraph{Robustness and adversarial vulnerability.}
Since the MEOW approach identifies the subset of units that carry the bulk of the representation's distinguishability, it also singles out the degrees of freedom on which the network's behaviour most sharply depends. This suggests using the ME as a lens on robustness: whether ME-critical neurons are disproportionately implicated in a network's sensitivity to adversarial perturbations and distribution shift, and whether ME-guided pruning or regularisation of these units alters the trade-off between accuracy and robustness. Such a study would tie the geometry of the hidden representation to the security properties of trained models.

\section*{Acknowledgments}

This work was conducted in the spirit of the Slow Science Manifesto, advocating for collaborative and sustainable research (slow-science.com).



\section*{Author contributions}

RP and AI proposed the study; RP, AI and MM conceived the work plan and proposed the method; AC and MM carried out the simulations; MM carried out the preliminary data analyses. All authors contributed to the analysis and interpretation of the data. MM drafted the manuscript. All authors reviewed the results, contributed to writing the paper, and approved the final version of the manuscript.

\bibliographystyle{ieeetr}
\bibliography{main}

\end{document}